\documentclass[manuscript,screen,nonacm]{acmart}

\setcopyright{none}
\renewcommand\footnotetextcopyrightpermission[1]{}
\usepackage{xr-hyper}
\usepackage{tabularx}
\usepackage{array}
\usepackage{booktabs}
\usepackage{tikz}
\usepackage{pdfpages}
\usepackage{pgfplots}
\usepackage{longtable}
\pgfplotsset{compat=1.18}
\AtBeginDocument{%
  }

\AtBeginDocument{%
  \hypersetup{
    colorlinks=true,
    citecolor=ACMDarkBlue,
    linkcolor=ACMDarkBlue,
    urlcolor=ACMDarkBlue
  }
}

\begin{document}

\title[Responsible GenAI Use During Armed Conflict: Evidence from Sudan's ERRs]{Responsible GenAI Use During Active Armed Conflict: Evidence from Sudan’s Emergency Response Rooms}




\author{Dalia Ali}
\authornote{Dalia Ali and Yearin Lee contributed equally to this work and share first authorship.}
\affiliation{%
  \institution{Technical University of Munich}
  \country{Germany}
}
\email{dalia.ali@tum.de}

\author{Yearin Lee}
\authornotemark[1]
\affiliation{%
  \institution{Technical University of Munich}
  \country{Germany}
}
\email{yearin.lee@tum.de}

\author{Cheng Yu}
\affiliation{%
  \institution{Technical University of Munich}
  \country{Germany}
}
\email{cheng.yu@tum.de}

\author{Ro Encarnación}
\affiliation{%
  \institution{University of Pennsylvania}
  \country{USA}
}

\author{Alsanosi Adam}
\affiliation{%
  \institution{Emergency Response Rooms}
  \country{Sudan}
}

\author{Danaé Metaxa}
\affiliation{%
  \institution{University of Pennsylvania}
  \country{USA}
}

\author{Ingmar Weber}
\affiliation{%
  \institution{Saarland University}
  \country{Germany}
}

\author{Orestis Papakyriakopoulos}
\affiliation{%
  \institution{Technical University of Munich}
  \country{Germany}
}
\email{orestis.p@tum.de}

\renewcommand{\shortauthors}{Ali \& Lee et al.}

\begin{abstract}

Armed conflict can increase the value of Generative AI (GenAI) while undermining the conditions required for its responsible use. Yet the perspectives of community-based humanitarian responders working under these conditions remain underrepresented in responsible AI research. We examine GenAI use in Sudan's Emergency Response Rooms (ERRs), a community-led humanitarian network, through an exploratory survey of 89 respondents and 15 semi-structured interviews with GenAI users and non-users. War reconfigures GenAI's role, making it most useful where its outputs are hardest to verify. ERR responders turned to GenAI when conflict made human expertise difficult or dangerous to reach. 
However, participants cautioned that GenAI could not independently establish rapidly changing local conditions, and that relying on its outputs could lead to harmful decisions when human verification was difficult. Risks arose before GenAI was used. For some volunteers, carrying a phone or accessing the internet could expose them to danger. Others were concerned that sharing sensitive beneficiary or operational information with GenAI could put people at risk. Participants feared that AI-assisted reports could undermine their credibility with donors, whose funding sustains the food and medical aid ERRs provide. Access to GenAI was uneven, and volunteers using it sometimes relied on colleagues with relevant expertise or local knowledge to verify its results. Our findings show that responsible use of GenAI during conflict depends not only on model performance or individual adoption, but also on whether responders can safely access these tools, verify their outputs against local conditions, and recognize when limiting or avoiding GenAI is the more responsible choice.

\end{abstract}




\maketitle



\section{Introduction}


Armed conflict affects civilians by causing harm, death, displacement, and limiting access to basic needs such as food, water, and health care. Its effects are widespread; in 2025, 117.8 million people were forcibly displaced, including 68.7 million displaced within their countries~\cite{icrc2026annual,unhcr2026figures}. In Sudan, since fighting started in April 2023, 9.3 million people have been displaced within the country, and 4.3 million have crossed into neighboring countries. In 2026, 33.7 million people there require urgent humanitarian assistance, and 37\% of health facilities are non-functional~\cite{who2026sudan}. International humanitarian organizations have faced access and capacity constraints, increasing the importance of locally led mutual aid like Emergency Response Rooms (ERRs), a community-led volunteer network that provides humanitarian assistance during the war in Sudan~\cite{carter2026supporting}. The term ``Rooms'' refers to the WhatsApp group chats through which volunteers initially organized their emergency response activities~\cite{k2024key}.

Under such conditions, Generative AI (GenAI), particularly large language models (LLMs), may provide a readily accessible source of information and assistance when human expertise or organizational support is difficult to reach. Previous research has documented the use of AI for crisis forecasting and resource allocation in humanitarian settings~\cite{bhatnagar2025bridging}. Similarly, practitioners in humanitarian, development, and environmental organizations report using LLM-based tools, such as ChatGPT and Copilot, for drafting and summarization~\cite{ali2025ai}. LLMs have also been used for humanitarian situation reports, which summarise the scope, impact, immediate needs, and ongoing relief efforts during crises, by synthesizing humanitarian documents into structured reports~\cite{decostanzi2026large}.



The use of digital technologies in humanitarian work can expand operational capabilities while introducing new operational, ethical, and political challenges~\cite{sandvik2014humanitarian}. Increased access to information does not automatically mean a commensurate ability to process and make use of it~\cite{read2016data}. In addition to these challenges, collecting and using digital data can itself create risks for populations affected by violence and insecurity~\cite{sandvik2017protective}. These tensions are particularly acute during armed conflict, where restrictions on humanitarian access can increase reliance on local responders who must operate in difficult and unsafe conditions~\cite{barter2023conflict}. 


Recent work has examined AI adoption, organizational readiness, and governance challenges in humanitarian contexts~\cite{bhatnagar2025bridging,attia2026responsible}, as well as AI use in humanitarian interventions~\cite{scuotto2025tackling} and LLM-based situation reporting~\cite{decostanzi2026large}.
However, much of this work focuses on formal humanitarian organizations, humanitarian professionals and experts, or systems introduced for specific organizational purposes. There is far less empirical knowledge about community-based responders' use of, limitations on, or avoidance of GenAI in their daily work during active armed conflicts. This gap matters because self-organized networks like Sudan's ERRs may lack the institutional protection and support available to larger humanitarian organizations. ERR volunteers also face disrupted infrastructure, limited access to expertise, and direct threats to their safety and the people they assist \cite{k2024key,hrw2024khartoum,amnesty2024sudan}. For the purposes of this study, we define responsible GenAI use as that which considers risks to responders and affected communities, protects sensitive information, and ensures appropriate human verification before outputs inform consequential decisions. 
We therefore ask:


\begin{itemize}
   \item[\textbf{RQ1:}] \textbf{How do community-based humanitarian responders use and/or avoid GenAI during active conflict?}
    \item[\textbf{RQ2:}] \textbf{How do wartime conditions shape GenAI accessibility, usefulness, and responsible use?}
\end{itemize}

To answer these questions, we study Sudan's ERRs, which rely on locally organized volunteers who work under varying infrastructural and security conditions, often without the institutional support available to formal humanitarian organizations. This network includes more than 700 ERRs across Sudan and over 26,000 responders~\cite{adam2026emergency}. We first conducted an exploratory survey with 89 ERR respondents to identify patterns of GenAI use and non-use, followed by 15 semi-structured interviews with ERR responders to examine these patterns in greater depth.

Our findings show that GenAI use in Sudan's ERRs is not a straightforward case of adoption or non-adoption. Some ERR responders turned to GenAI when conflict made human assistance difficult to reach, but they also recognized its limitations in interpreting rapidly changing conditions on the ground. Safety concerns shaped both whether GenAI could be used and what information could be shared with it, while GenAI-assisted humanitarian communication introduced questions about accuracy, representation, and credibility. Moreover, access to GenAI and responsibility for reviewing its outputs were distributed across responders operating under unequal infrastructural and security conditions. In total, we contribute the following:

\begin{enumerate}
\item \textbf{Empirical Evidence on GenAI Use During Active Armed Conflict.}
We provide new empirical evidence on the ways community-based humanitarian responders use, limit, and avoid GenAI during active conflict, bringing the perspectives of volunteers delivering essential humanitarian assistance under direct security threats, disrupted infrastructure, and limited access to expertise into responsible AI research.

\item \textbf{Conflict-Specific Conditions of Responsible GenAI Use.}
We show that active conflict can simultaneously increase the value of GenAI and undermine the conditions required for its responsible use, a tension starkest in life-or-death situations.  
For example, a responder unable to safely access medical care might find great value in GenAI assistance, but simultaneously be put at risk by carrying the phone needed to access it.
Responsible GenAI use therefore depends not only on the model reliability and individual use, but also on situated human knowledge, task-specific boundaries, and how responders coordinate access to GenAI and review its outputs.

\item \textbf{Conflict-Sensitive Design and Governance of GenAI.}
We derive implications for humanitarian organizations, donors, and technology providers that address safety risks before and during GenAI use, preserve human judgment and the legitimacy of non-use, and safeguard the credibility and accountability of AI-assisted humanitarian work.
\end{enumerate}

\section{Related Work}
This section reviews current applications of AI and GenAI in humanitarian response. It also addresses challenges associated with GenAI use in high-stakes contexts, identifying gaps in existing work that our study aims to address.

\subsection{Use of AI and GenAI in Humanitarian Response}

Armed conflicts, natural disasters, and political pressures disrupt the delivery of essential humanitarian services to affected communities~\cite{bradley2023politics,akpoghelie2024malnutrition,csenol2025emergency}. Under such conditions, AI has demonstrated its utility across different phases of crisis response~\cite{LYTHREATIS2026103415}. 
Before crises occur or escalate, AI can support early warning and anticipatory planning. AI can predict some crises using satellite imagery, household surveys, mobile phone records, and social media data~\cite{he2026artificial}. For instance, HungerMap LIVE uses machine learning (ML) to monitor food insecurity in near-real time across more than 50 countries and provide actionable insights into where food insecurity is most severe~\cite{vos2023food,wfp2026hungermap}.

During crises, AI has also been used to support situational awareness, emergency coordination, and damage assessment in affected areas. For example, the Qatar Computing Research Institute developed AIDR and tested it during the 2013 earthquake in Pakistan, classifying disaster-related social media messages using ML and providing actionable information in real time~\cite{imran2014aidr}. Similarly, World Food Programme's SKAI tool, developed with Google Research, uses ML to detect damaged infrastructure from satellite imagery, with reported uses during Hurricane Ian and the Pakistan floods in 2022, and the Türkiye earthquake in 2023 
~\cite{itu2021united,fomina2026artificial}. Recent studies have explored the potential of LLMs to analyze and synthesize disaster-related information and generate reports from the data for emergency response planning and coordination~\cite{xu2025large}. AI can also contribute to post-crisis recovery efforts by facilitating rapid damage assessments and identifying priority areas for recovery. The United Nations Development Programme's RAPIDA tool has been applied for this purpose in response to the 2026 earthquakes in Venezuela and cyclones in Madagascar~\cite{qehaja2026crisis}.

However, these applications are often dependent on technical infrastructure, expertise, and organizational resources that may be challenging for decentralized local humanitarian organizations to secure. 
Commercially available GenAI tools like ChatGPT could provide an alternative source of necessary support, since humanitarian responders can more easily access them, without developing dedicated AI systems. Such tools may, however, pose significant risks, as they are not explicitly designed for conflict environments. Limited research has examined how local humanitarian responders use, limit, or avoid GenAI in their work, particularly during active conflict, or documented these practices in their own voices. 
We examine these practices among ERR volunteers in Sudan, providing empirical information from community-based responders whose experiences are underrepresented in responsible AI research.

\subsection{Sociotechnical Risks of GenAI in High-Stakes Contexts}

The use of GenAI in humanitarian response requires considerations beyond practical challenges, as its limitations can lead to real-world harm. One concern is bias, which can emerge from unrepresentative training data and be reinforced during model inference, potentially disadvantaging certain populations~\cite{afreen2025systematic}. 
Many language technologies have also been shown to perform unevenly across languages or fail to represent linguistic and cultural specificities~\cite{bella2024tackling}. Privacy presents another challenge, as LLMs can memorize and reproduce personal information from training data~\cite{carlini2021extracting}, which can pose considerable security risks in sensitive contexts. LLMs can also infer sensitive personal attributes such as location and income from contextual details, even when explicit identifiers have been removed~\cite{staab2024beyond}.


Another major concern is hallucination, which occurs when LLMs generate factually conflicting or unverifiable information or produce responses that are inconsistent with user instructions, provided context, or their own reasoning~\cite{huang2025survey}. 
For example, news outlet CNN reported that an AI-assisted US military intelligence report falsely identified a Chinese vessel as transporting nuclear weapons components, prompting preparations for a military interception before the error was detected~\cite{lillis2026military}. 
GenAI can also be misused to produce misinformation and disinformation, particularly through deepfakes~\cite{uddagiri2024ethical}. For example, during Russia's invasion of Ukraine in 2022, a deepfake video falsely depicted Ukrainian President Volodymyr Zelensky calling on Ukrainians to surrender to Russia~\cite{wakefield2022deepfake}. 

Difficulties in accurately assessing conditions on the ground can further complicate effective humanitarian response. For example, a review of AI-based food insecurity prediction identifies the lack of ground-truth data in conflict settings as a barrier to accurately assessing food insecurity~\cite{he2026artificial}. Another study found substantial differences in how human annotators labeled damage to the same buildings when using drone versus satellite imagery and raised concerns about potential risks of AI damage assessment systems trained on such imagery-derived labels~\cite{manzini2025now}. Related reliability challenges have also been observed in GenAI. Research has shown that LLMs often perform poorly in some low-resource languages used in crisis settings and that an LLM assigned markedly different urgency levels to the same simulated crisis scenarios presented in different languages~\cite{ticona2026llm}. While these studies discuss important limitations in data quality and model performance, less attention has been paid to how humanitarian responders encounter and manage them in practice. We address this gap by examining how ERR volunteers in Sudan perceive GenAI's risks, verify its outputs, and decide when to limit or avoid its use during active conflict.

\section{Method}
The study comprised a two-stage design. We first conducted an exploratory survey to develop an initial understanding of GenAI use and non-use among ERR responders and to identify issues for deeper investigation. We then conducted semi-structured interviews to examine these issues and participants' experiences in greater depth.

\subsection{Ethics and Participant Safety}

Given the sensitivity of conducting research with humanitarian responders during active armed conflict, approaching this work with extreme care and appropriate positionality was crucial. Two members of the research team are part of the ERR network and are familiar with the Sudanese humanitarian context. Their background knowledge and existing relationships within the network informed the study design, participant recruitment, and data interpretation.

We obtained Institutional Review Board (IRB) approval and took additional steps to address participant safety, confidentiality, and barriers to participation in the context of the ongoing war. Specifically, participants were encouraged to avoid providing names, precise locations, or operationally sensitive information, and we removed identifying details during data analysis and reporting. To reduce barriers to participation, we offered to cover participants' internet costs. 

After conducting the interviews and before processing the data, we assessed whether combinations of location, role, and experience could identify participants within the ERR network or expose them to harm, and generalized details where necessary. Interview recordings, transcripts, and survey data were stored privately, with access limited to researchers involved in the analysis. For the LLM-assisted translation and coding, which participants were informed about verbally before recording, we used GPT-6 through a ChatGPT Pro account provided by the research group~~\cite{openai2026gpt6} (details in Appendix~\ref{app:appendixB}). We only provided de-identified documents to the model, after removing or generalizing information that could directly or indirectly identify participants; we disabled model-improvement data sharing before submitting any content to ChatGPT.

Participants could choose not to answer particular questions or stop the interview at any time, as well as request withdrawal from the study after the interviews were conducted. Moreover, because interviews might include discussions of events experienced in the context of conflict that could cause emotional distress, the study protocol also included access to mental-health support or referral information if participants requested it (although none did).

\subsection{Stage 1: Exploratory Survey}

We started by conducting an exploratory survey with ERR responders to identify issues for deeper investigation in the interviews. The survey was designed and translated into Arabic by a native Arabic-speaking researcher and was open to ERR responders aged 18 years or older. Survey respondents were recruited through network-based and participant-referral sampling. A research team member who is part of the ERR network initially distributed the survey to existing ERR contacts, who subsequently shared it with other volunteers. In some cases, responders with internet access also reached those without connectivity and provided access to the survey so that they could complete it in person.

The survey focused on the roles and humanitarian activities of the participants, their experience with the ERRs, their use and non-use of GenAI, the tools they use, the tasks they perform, the languages and devices used, the reasons why they do not use GenAI, and how they verify AI outputs. Moreover, there were 23 Likert-scale questions on technological, organizational, and environmental aspects, including access to devices and the internet, organizational support and training, privacy and security issues, and the necessity of human verification.

The final survey sample consisted of 89 unique respondents after removing duplicate submissions with identical response information. The survey was exploratory and was not intended to provide representative estimates of GenAI use across ERRs. We used the survey responses descriptively to characterize the sample and identify initial patterns that informed the design of the semi-structured interview guide, including questions about output verification, sensitive information, and internet access. The full questionnaire and further descriptive statistics are provided in Appendix~\ref{app:appendixA}.

\subsection{Stage 2: Semi-Structured Interviews} 
Following the exploratory survey, we invited some respondents who had agreed to be contacted for follow-up interviews, which were conducted between August and September 2026. Among respondents who agreed to be contacted and provided contact information, we purposively selected both GenAI users and non-users. We also sought variation in geographic areas and roles within the ERRs, including participants working in areas under the control of different armed actors. We conducted 15 semi-structured online interviews with ERR volunteers, each lasting approximately 45--60 minutes. The final sample included nine participants who had used GenAI in their ERR work and six who had not. We allowed participants to choose the language they were most comfortable with. We therefore conducted 14 interviews in Arabic and one in English.

For GenAI users, we asked about their experiences with GenAI, including the circumstances that led them to use it, the data entered, how outputs were used and checked, and situations where participants chose not to use GenAI or limited its use. For non-users, we explored indirect exposure to GenAI, reasons for non-use, existing alternatives, and the conditions under which they might consider using it. Both paths examined how the ongoing war affected access, safety, and decisions about appropriate use.

The guide included a sector-specific question based on each participant's work, such as health, food distribution, protection, logistics, information management, or donor relations. These questions examined whether GenAI was used in potentially high-risk activities, what participants considered appropriate or inappropriate, and where human decision-making was important. With participants' permission, interviews were recorded for transcription. Participants using GenAI could share their screen to demonstrate their workflow, but were instructed not to display identifiable, confidential, or operationally sensitive information. Interview guides are provided in Appendix~\ref{app:appendixB}.


\vspace{-0.18cm}

\subsection{Data Analysis}

We used the survey responses descriptively to characterize the sample and identify initial patterns to inform the subsequent interviews. A native Arabic-speaking author transcribed the 14 Arabic interviews and translated them into English with AI assistance, manually reviewing the machine translations against the Arabic originals. 

We analyzed the interview data using a codebook approach to thematic analysis~\cite {braun2006thematic,braun2021one}. Our coding was primarily inductive, with codes developed from participants' responses instead of from a predefined theoretical codebook. We first read the transcripts and coded relevant segments in QCAmap~\cite{fenzl2017qcamap}, a qualitative data analysis tool; then we used GPT-6 to generate an additional set of candidate codes and possible patterns for researcher review. We did not treat the LLM as an independent coder, and we checked all suggested codes and patterns against the interview transcripts and our own judgment. As coding progressed, we compared both sets of codes with the transcripts, compared similar codes, revised their definitions, and removed unsupported or redundant codes. 

We then conducted a collaborative analysis workshop involving four members of the research team to map related codes and develop higher-level themes. During the workshop, we compared GenAI users and non-users and examined patterns across different forms of humanitarian work. We grouped related codes, discussed differences and recurring patterns, and returned to the transcripts to check emerging themes against the interview data. More details on coding and thematic analysis are in Appendix~\ref{app:appendixB}.

\begin{table}[h]
\centering
\caption{Overview of semi-structured interview participants ($N=15$).}
\label{tab:interview-participants}

\footnotesize
\setlength{\tabcolsep}{3.5pt}
\renewcommand{\arraystretch}{1.02}

\begin{tabular}{
@{}
p{0.055\columnwidth}
p{0.18\columnwidth}
p{0.3\columnwidth}
p{0.36\columnwidth}
@{}
}
\toprule
\textbf{ID} &
\textbf{Location} &
\textbf{Working area} &
\textbf{Role / main focus} \\
\midrule

\multicolumn{4}{l}{\textit{\textbf{GenAI users ($n=9$)}}} \\

P01 & National &
External communications / advocacy &
External liaison and advocacy \\

P02 & Red Sea &
Finance &
Financial reporting, transfers, and suppliers \\

P03 & Khartoum State &
Monitoring and evaluation &
Implementation and quality monitoring \\

P04 & Khartoum State &
External relations &
Fundraising, partnerships, and advocacy \\

P05 & Gezira / National &
Programs coordination &
Multi-level program coordination \\

P06 & Khartoum / National &
Protection / gender &
Women's programming and safe spaces \\

P07 & Khartoum &
Logistics / food assistance &
Community kitchens, procurement, and supplies \\

P08 & Khartoum &
Health / medical coordination &
Medical coordination and health projects \\

P09 & South Darfur &
Programs / project management &
Needs assessment and project implementation \\

\midrule
\multicolumn{4}{l}{\textit{\textbf{GenAI non-users ($n=6$)}}} \\

P10 & West Darfur &
Programs / project management &
Planning, implementation, and reporting \\

P11 & National &
Communications / advocacy &
Field information, donor liaison, and advocacy \\

P12 & Khartoum &
Emergency / medical relief &
Hospital support and emergency medical relief \\

P13 & River Nile &
Community relief &
Kitchens, distributions, and basic services \\

P14 & South Kordofan &
Communications / field coordination &
Donor, volunteer, and distribution coordination \\

P15 & North Darfur &
Programs / needs assessment &
Project planning and community coordination \\

\bottomrule
\end{tabular}

\vspace{3pt}
\parbox{\columnwidth}{
\footnotesize\textit{Note.} ``National'' refers to ERR work spanning multiple states.
Combined labels such as ``Khartoum/National'' indicate work based in the named
state with additional national-level responsibilities. Location and role
descriptors were reviewed by the research team, including a researcher who is
part of the ERR network, and by an additional person familiar with the ERR
network, to assess whether combinations could identify participants or expose
them to harm.
}

\end{table}


\section{Results}
\label{res:result}
Of the 89 respondents, 49 (55.1\%) reported using GenAI in their ERR work, 30 (33.7\%) reported not using it, and 10 (11.2\%) were unsure. Among respondents, 65.2\% agreed that GenAI could be useful in fast-paced ERR work, while 70.8\% agreed that its outputs usually required verification, and only 41.6\% agreed that GenAI-generated information was usually accurate. Half of respondents (50.6\%) expressed concern about exposing sensitive information and only 44.9\% reported sufficiently stable internet access. These patterns suggested that perceived usefulness of GenAI coexisted with concerns about its reliability, information protection, and accessibility. Additional survey results are in Appendix~\ref{app:appendixA}.

\subsection{Turning to GenAI When War Makes Human Support Hard to Reach}
GenAI became useful for some ERR responders because active fighting had made ordinary routes to human assistance unsafe, slow, or unavailable. Participants described scenarios in which consulting a doctor, reaching a specialist, or returning to a site with the right equipment was no longer straightforward. 
P07 illustrated this with a hypothetical situation in which health emergencies could arise among people sheltering during the fighting:

\begin{quote}
``One of the girls might have a seizure. You do not have a doctor at hand, and taking her to a doctor would itself put her life at risk. You have to turn to [GenAI]. [...] If no one around you can help solve the problem, the quickest thing is to use [GenAI].'' (P07)
\end{quote}

Beyond food, housing, and medical care, P07 described psychological needs among people sheltering during the fighting, like some people becoming so frightened that they refused to eat or drink, and children being exposed to \textit{``gunfire or the sound of gunfire.''} P07 used GenAI chatbots to generate suggestions for things they could do in such moments, such as activities for children and ways to move people away from places where the sounds of fighting were most distressing. 


A similar pattern appeared in a technical assignment; P03 recounted how Gemini helped him during the analysis of a solar energy system installed at a dialysis center, including assessing the risk of shadows on the solar panels based on photographs and the position of the sun. In the moment, he was faced with an unforeseen challenge, lacked the proper equipment, and was uncertain about his ability to revisit the site later due to the active war:
``First, I did not have a measuring tape. Second, there was nobody with me, and I did not think I would be able to go back again, so I used Gemini straight away.'' (P03)

P03 treated Gemini's output as an initial assessment and discussed it with the engineer responsible for the installation, who said he had already accounted for the issue and would address it. In this case, GenAI provided provisional assistance rather than replacing human expertise. While P07 described why immediate GenAI assistance could be valuable when reaching professional help was delayed or impossible, P03's experience demonstrates how the need for additional information could prompt its use even in the presence of a human expert.
These accounts show how wartime conditions could move GenAI into situations where responders would otherwise seek human support.

\subsection{GenAI Could Not Independently Establish Wartime Ground Truth}

Participants distinguished between information that they were confident GenAI could process and that which it could not, including the current circumstances needed to make humanitarian decisions. During active war, circumstances could quickly change after information had been collected or entered into the system. Prices, currency exchange rates, weather, population migration, and resources could change in days. Such fluctuations were highlighted by P04 when talking about using GenAI in project planning:

\begin{quote}

``Today, you might design a project that costs 600, 800, or 1,000 dollars. Within a week, by the time you implement it, things have changed. The market is not stable; the exchange rate is not stable [...] [GenAI] only gives you indicators that you can build on. But in the end, the final decision should be made by the human who is on the ground and sees how things are actually going.'' (P04)
\end{quote}

The need for up-to-date knowledge of the locality also influenced how willing the participants were to use GenAI. P08 utilized GenAI for organizing, but set limits to its usage in initiating projects or determining what the community needs. Project development instead began with information obtained directly from volunteers and affected communities. When asked why she would not use AI to develop the core idea for a project, she was direct: \textit{``AI does not know the actual situation. Instead, we turn to the volunteers themselves.'' }(P08) For her, GenAI could help organize what responders already knew, but it could not replace the local knowledge used to decide what should be done. 

P11, a non-user, made a similar distinction between information available to a model and knowledge attained from living and working in the affected environment. He described local problem-solving as an iterative process in which responders tried an approach and adjusted it as conditions like weather, materials, or community needs changed. He argued that online information could not capture people's daily struggles or the repeated adjustments involved in responding in times of war:

\begin{quote}

``[GenAI tools] don't know about the struggle, the daily life of the people. What do you know about what the need is and how to respond to it? The people know, right? The people on the ground. It's not a one-question, one-answer situation.'' (P11)

\end{quote}

These cases show different limits to relying on GenAI-generated information during the war. P04 emphasized that GenAI-generated cost estimates could become outdated as market prices and exchange rates changed, affecting project planning and implementation. P08 and P11 pointed to a different limitation, as identifying community needs and adapting humanitarian responses required knowledge developed through direct engagement with affected people and changing circumstances. Notably, this concern appeared in both P08's selective use of GenAI and P11's non-use. For these participants, evaluating whether GenAI-generated information could inform humanitarian decisions required more than assessing the output itself; it also depended on current knowledge of the situation on the ground. Consistent with this, only 49.4\% of survey respondents agreed that GenAI understands local words, places, and context in Sudan.

\subsection{Wartime Safety Concerns Shape GenAI Use and Information Disclosure}

For some responders, the safety concerns related to technology use impacted whether GenAI could be used at all. P12 described the period during which he was operating on the front lines, got arrested, and often either hid or discarded his phone: ``I always left my phone behind at the hospital or hid it somewhere. Even when I was arrested, thank God, they did not find my phone with me. At that time, I was not thinking about AI.'' (P12) Later, P12 described that in this period ``having a phone was itself a crime''. Armed actors could search phones and interpret messages, search histories, or social media activity as support for the opposing side, potentially exposing their owners to detention or death. 

Among participants who did use GenAI, the same safety issues determined what kind of information they permitted it to access. P06 engaged with data related to women who have accessed help services or participated in events. She was comfortable storing such information digitally, but drew a clear boundary around giving the same data to GenAI. She also gave the example of handwritten beneficiary lists that could be converted into a digital document using AI:

\begin{quote}
``Those are not something we would upload [to an AI tool]. We put them in Drive [...] I do not think I would ever take one of those lists, for example, a handwritten one that I want converted into a PDF or a document, and put it into an AI tool and say, `Convert this for me.' Those really are sensitive data.'' (P06)
\end{quote}

P06 saw a distinction between digital storage and disclosure to GenAI. She described GenAI services as ``open websites'' where she could not be confident that sensitive information would remain securely stored. When discussing the greatest risks, she specifically referred to account numbers, beneficiary lists, and volunteer incentive payments as information that should not be uploaded. 

P09 described a different way of setting this boundary while using GenAI for project and reporting work. Rather than avoiding the technology, he separated the parts of a task he was willing to give to the system from the information he kept outside it. He added names, budgets, and other details manually after receiving output text from GenAI. 
P09 explained that he avoided providing personal and operational information because he worried that it could be used for purposes beyond those he intended or become exposed.

\begin{quote}

``Preparing budgets, numbers, and those things, I never enter them into AI. I only ask it to write an overview of the project, the project's objectives, and so on. [...] I do not tell AI that this speaker was so-and-so from such-and-such a place.'' (P09)
\end{quote}

However, three other participants who stated similar boundaries also described workflows that appeared to involve uploading names, numbers, or original records, although their accounts did not establish whether this information had been de-identified.

In these cases, the wartime safety concerns influenced the use of GenAI at different points. For P12, the danger associated with digital participation led to the non-use of GenAI during the period when he was in danger. For P06 and P09, the question instead was about what information could be disclosed to GenAI while still using the technology. P06 was concerned about losing control over sensitive beneficiary and financial information, while P09 described possible consequences for other volunteers and the wider ERR if information became exposed. These show that wartime safety concerns shaped both whether responders could engage with GenAI and how they limited its involvement in humanitarian work.

\subsection{GenAI Complicates the Credibility of Humanitarian Communications}

Participants expressed concerns about using GenAI to communicate humanitarian efforts to donors and affected communities. Credibility in such cases required not only the clarity of the generated text and the ability to present it in a grammatically correct way, but also that reports and public communication correspond to what had actually happened, use language that people could understand, and remain recognizable as the work of the people producing it. If donor doubts about GenAI-assisted reporting affected funding, ERRs could struggle to provide food and medical assistance, with potentially life-threatening consequences for the communities they serve.

P10 was particularly concerned about GenAI-generated reporting becoming detached from activities on the ground. He described reporting as an account of what had actually been implemented, including its effects, challenges, and feedback from beneficiaries:

\begin{quote}

``It is a matter of credibility; you need to convey things as they happened, particularly in the program work concerning the impacts of the activities. The lessons learned, the challenges. The impact on the beneficiaries. ChatGPT might give you information different from what you want, and it might give you information you would not expect to have been there on the ground. So it becomes a matter of the credibility of the work.'' (P10)

\end{quote}



Participants also raised questions about whether fluent AI-generated text was appropriate for its intended audience. P15, a non-user, explained that humanitarian communication needs to include local cultures, customs, and languages, including communities for whom Arabic might not be readily understood. His account identifies a requirement against which GenAI-generated communication would need to be assessed. In this case, appropriate communication could not be judged from the fluency of the generated text alone; it required knowledge of the people receiving it.

\begin{quote}
``There also has to be consideration for people's cultures, their customs and traditions, and the language communicated to people should also match what they can understand [...] The communities now in the northwest, bordering Chad, most of them -even if you speak Arabic-, a large proportion do not understand what you are saying at all, unless it is explained to them in their own local language.'' (P15)
\end{quote}

Credibility could also be affected even when participants believed that AI had only played a limited role. P05 described a project proposal that his team had written themselves and then given to AI to organize and turn parts of the material into tables. A donor subsequently questioned whether the entire proposal had been generated by AI:

\begin{quote}

``We wrote it entirely ourselves and gave it to [AI] only to organize it, nothing more. It made tables for us. Then the donors spoke to us and said that they thought the whole project had been written by AI [...] But we had written it all ourselves. We had only used AI to arrange or organize it.'' (P05)

\end{quote}

Although P05's example does not prove that the donor detected the use of AI correctly or that the proposal was rejected, it does show that even limited GenAI assistance could create uncertainty about authorship and the legitimacy of a humanitarian document.

All these examples demonstrate that the concern was not simply whether GenAI produced readable text. Humanitarian communication needs to remain connected to the activities being reported, relevant to the audience it is delivered to, and credible to external partners. GenAI could help with writing, translating, organizing, and presenting texts, but its use also entered existing relationships between responders, affected communities, and donors. Participants therefore viewed human judgment as necessary not only for reviewing what GenAI produced, but also for determining whether the resulting communication appropriately represented their work and the people involved.

\subsection{Unequal Access and Collective Use of GenAI During Wartime}

The use of GenAI by ERR responders depended not only on personal willingness, but also on differences in security, connectivity, access to devices, paid services, and educational opportunities. At the same time, GenAI use was shaped by the distribution of work among volunteers, including who had access to these tools, who held relevant local knowledge, and who could review their outputs.

P13, a non-user, believed that there was no urgent need for GenAI since reporting and data work were already handled by another person in his emergency response room: 
``We do not need [GenAI tools]. In our room someone specializes in data and works on it, communicates with the rooms, prepares the reports and submits them.'' (P13)

P14 described a different reason for non-use. Her work focused on organizing volunteers and distributions through direct interaction with people and was largely field-based. She had never used GenAI and also reported limited access to smartphones and reliable connectivity, although she expressed interest in using it if appropriate devices and training became available. For both P13 and P14, non-use did not necessarily indicate rejection of GenAI.

P15 anticipated that responders in safer locations would have better opportunities for learning and using GenAI than responders in less secure areas, describing a concern about a possible future difference in GenAI experience rather than an observed difference in proficiency:

\begin{quote}
``I think people in safe places will get ahead of those in unsafe places in how to use artificial intelligence and benefit from it. They will keep advancing if the war does not stop [...] If we had the same space that they have, we would use artificial intelligence.'' (P15)
\end{quote}

Even among GenAI users, access was unstable. P02 described electricity cuts and disconnected networks that postponed tasks and created backlogs. Poor and expensive connectivity additionally limited GenAI use. Other participants described that access to paid GenAI was available through certain projects or offices, user accounts were shared among colleagues, and better-connected people were taking care of the tasks that were not available to others. GenAI access in such a way depended on the resources within the ERR rather than solely on resources held by an individual responder.

Access to the tool did not remove dependence on human review. P02, who used GenAI in financial work, argued that checking should reflect the consequences of the task:

\begin{quote}
``We use [GenAI] to help us, but the output needs to be reviewed. There needs to be a human element going through it and reviewing it [...] With financial matters, you need to check even the fraction, tiniest detail [...] The person generates something without going and verifying it and checking its accuracy; that is entirely their responsibility. We cannot leave it at `It was the AI's fault.'~'' (P02)
\end{quote}

P01 described a similar process in which GenAI outputs could be reviewed by the individual user, the wider team, or another responsible person before being acted upon. He also argued that responders should develop the skills needed to perform the task independently before relying on GenAI, warning that otherwise ``instead of it being your assistant, you become its assistant.'' (P01)

One responder may hold the field knowledge, while another may have better access to the device or network needed to use GenAI, and another may be responsible for checking and reviewing the output before it is shared or acted upon. In this sense, GenAI engagement cannot be understood only at the level of the individual user, since its use also depends on task roles, access to infrastructure, and human review.

\section{Discussion}
Our findings show that active conflict can increase the value of GenAI for community-based responders while undermining the conditions required for its responsible use. We first discuss how this tension shapes GenAI use, then derive implications for humanitarian organizations, donors, and technology providers.

\subsection{War Reconfigures the Value and Limits of GenAI}

Previous work has shown that AI can support humanitarian efforts while simultaneously raising concerns about inaccurate outputs, data privacy, governance, and organizational readiness~\cite{beduschi2022harnessing,bhatnagar2025bridging}. Our findings extend this work by showing how active conflict changes both the circumstances that make GenAI assistance valuable and the conditions under which humanitarian responders can access and use it. ERR responders turned to GenAI when it was difficult or dangerous to contact doctors, technical experts, and colleagues. However, war conditions made electricity, the internet, and paid services less accessible, restricted the availability of devices, and decreased the opportunities to learn. 
When expert human assistance was difficult to reach, GenAI could offer temporary support, but the same lack of access could delay the qualified human review needed to assess that support. This creates a tension for responsible GenAI use during wartime, insofar as the circumstances in which GenAI becomes most useful can also be those in which appropriate human verification becomes more difficult.

The observation that GenAI is not able to independently verify ground truth in wartime adds another limitation. Studies on conflict forecasting show that recent external information and retrieval can improve LLM performance~\cite{nemkova2025large}, while work on humanitarian reporting shows that retrieval from verified sources and source-linked citations can ground generated outputs~\cite{decostanzi2026large}. Our participants pointed out changes in prices, needs, population movement, available resources, and security conditions throughout the humanitarian response process. As a result, relevant information may not yet exist in any documents available to be retrieved by a model, particularly when conditions can change faster than they can be documented. This is not a problem of information recency alone. Part of the knowledge needed for humanitarian action is created through interaction with affected communities, local volunteers, and changing conditions on the ground~\cite{finn2016fundamentally,widder2024epistemic}. 
While GenAI can analyze the available information, it is incapable of replacing local human activities through which some of that knowledge is established and updated.

The context of active conflict also reshapes the consequences of GenAI's limitations. Research on crisis scenarios has shown that LLMs can assign different levels of urgency to the same situation depending on the language used~\cite{ticona2026llm}. Similarly, a benchmark of LLMs for coding humanitarian data found that aggregate performance measures can obscure weaknesses in identifying concerns related to physical safety and discrimination~\cite{marston2026can}.
Our participants treated fluent output as insufficient when it did not match the local language, the current situation, or the experience of the community. Our findings further show that these safety concerns can arise before any information is entered into GenAI. In some situations, having the device or going online was risky.
Responsible GenAI use in active conflict therefore cannot be assessed only through model outputs or the data submitted to the system; it also depends on the conditions under which people access the technology and the wider consequences that may follow from its use.

War may also shape the opportunities people have for developing GenAI competence over time. P15 anticipated that responders in a safer environment would have more opportunities for learning and becoming familiar with GenAI, while responders facing higher levels of insecurity might have fewer opportunities. This corresponds to the literature, which associates humanitarian AI adoption with the resource constraints of organizations~\cite{bhatnagar2025bridging}, as well as to literature that discusses the role of power inequalities in participation in humanitarian AI~\cite{suge2026experimentation}. 
Yet even where responders had access to GenAI, they did not necessarily have the expertise needed to verify its outputs. Within ERRs, the responder using GenAI was not always the person with current local knowledge or responsibility for reviewing its outputs. Responsible use therefore depended on both access to the technology and coordination with responders who could assess the information it produced.

\subsection{Implications for Conflict-Sensitive GenAI and Humanitarian Practice}

\subsubsection{Safety before and beyond GenAI use.} Our findings indicate that in an active conflict setting, the risks associated with GenAI could emerge even before any information is fed into the model. The use of GenAI might involve carrying a device, accessing internet connectivity, or digitizing information at a point where such digital activities raise safety concerns. Therefore, it is important for humanitarian organizations and ERR coordinators to establish a protocol for assessing whether to use GenAI for specific tasks given particular security circumstances. The questions to be asked include whether the use of technology puts a responder in additional danger, whether performing the task requires the sharing of sensitive humanitarian information, and whether such disclosure could create risks for other people involved in the humanitarian response.

If such concerns apply, responders might have to limit the use of GenAI or perform tasks without it. Where GenAI use is considered appropriate, responders could remove identifying or sensitive information before submission, although this may not eliminate all disclosure risks. The urgency of the situation does not exempt consideration of information sensitivity. Input from people familiar with the situation on the ground is important for such decisions, with additional specialist consultation where feasible, because general rules do not take into account possible changes in local security. Alternatives to the use of GenAI require a similar evaluation, because transferring the task to another responder may shift the risk rather than remove it.

Technology providers should enable such decision-making by clearly indicating whether submitted information is being stored, for how long, and for what purposes. Users should have practical ways to limit the retention of sensitive humanitarian information, including zero-data-retention options where feasible. All users deserve these abilities, but heightened risks make them especially high-stakes in active conflict settings.

\subsubsection{Verification when current information and expertise are unavailable.} Our findings indicate that GenAI can be especially useful in responding to war situations when reaching doctors, engineers, or other experts becomes challenging, though these conditions also make its outputs harder to verify. Information about prices, resource availability, needs, or local conditions can change quickly, become out of date, or be hard to verify safely. Humanitarian organizations and ERR coordinators should set task-specific boundaries on GenAI usage. Such boundaries should distinguish cases where GenAI can provide provisional assistance for tasks where errors have limited consequences from cases where its output requires qualified human verification before action is taken.

Where possible and necessary, ERR coordinators could establish peer-review procedures involving responders with the expertise and current local knowledge required to verify the output, while considering whether sharing information for verification creates additional risks. In the absence of verification and when immediate action is necessary, GenAI may still provide provisional information, but its unverified output should not become the sole basis for consequential decisions. Organizations should retain humanitarian procedures or local knowledge as alternative sources for decision-making when expert review is unavailable.

Technology providers should facilitate assessment of whether the information obtained by a user is current by providing clear model knowledge cutoff dates and, where information is retrieved, verifiable sources and dates of retrieval. These elements would help establish the origin and timeliness of information, even if they cannot guarantee that it corresponds to the situation in the area. 

\subsubsection{Unequal access and the legitimacy of non-use.} Our findings suggest that opportunities to access GenAI tools and acquire relevant skills may differ depending on the region and security circumstances of responders; those in safer and better-connected areas may have greater opportunities to access paid tools or knowledge about GenAI. At the same time, insecurity, low connectivity, few devices, and urgent humanitarian work may limit such opportunities. Moreover, non-use can be deliberate, as some volunteers prefer not to use GenAI for certain tasks because of security concerns or a preference for human judgment. Non-use can also occur when GenAI is not relevant to a responder's responsibilities or when another person already performs the relevant task.

Therefore, humanitarian organizations and ERR coordinators should differentiate between non-use caused by access constraints and cases where GenAI is unnecessary or deliberately avoided. Training should be voluntary and related to responders' responsibilities, while those who choose or need to use GenAI should be supported. Where funders and organizational leadership support GenAI use, they should allocate dedicated funds where possible and avoid initiatives that divert resources from emergency humanitarian operations, since responders reported prioritizing humanitarian needs over GenAI spending. Shared accounts and devices may improve access to GenAI but can expose sensitive information or make responsibility for submitted material unclear. Technology providers could consider offering lower-cost humanitarian access, low-bandwidth interfaces, and lightweight models suited to unreliable connectivity. 

\subsubsection{Credibility and representation in humanitarian communication.} Participants highlighted concerns related to the possibility that a report prepared with GenAI or a summary of a discussion could misrepresent the situation on the ground. At the same time, volunteers use GenAI to write English-language documents for mainly English-speaking funders. This creates a tension, since the same technology that can improve communication can also distort the message or cast doubt on document authorship.

Humanitarian organizations and ERR coordinators should connect GenAI-assisted reports with evidence of the activities and check translations and documents with the involvement of people who know the local context. Where possible, this evidence should support the claims made in the document without requiring the disclosure of sensitive operational information. Donors should distinguish between the use of GenAI for translation, editing, or organizing documents and the fabrication of claims, and assess documents based on evidence supporting their claims rather than assumptions about AI authorship alone, particularly when volunteers write in a non-native language. They should also consider the time-sensitive and risky conditions under which volunteers work, and the need for rapid assistance from GenAI when human assistance is difficult to obtain. Technology providers could provide optional ways of distinguishing user-written information, information added by GenAI, and human changes while keeping sensitive operational information undisclosed.


\subsection{Limitations}

There are limitations to this research. First, the sample of the exploratory survey and interview is not representative of the diversity of ERRs in Sudan. Darfur and Kordofan were less represented among interviewees than in the survey, and participation largely required connectivity, devices, and relative safety, likely underrepresenting the most constrained responders. 
In addition, recruitment through existing ERR contacts and participant-referral sampling could have influenced the composition of the sample and how participants reported their experiences. The geographic distribution of users and non-users also differed, limiting our ability to distinguish use-related patterns from regional conditions.

The interviews mostly depend on participants' reports on the usage of GenAI. We invited participants to share their screens, but all but one declined to do so due to security and privacy concerns. Thus, we were unable to directly observe the usage of GenAI and verify its effectiveness. Medical, technical, and humanitarian cases thus need to be interpreted as participants' reported experiences, rather than evidence of effective GenAI performance. 
We encourage future work examining the applicability of these findings to other conflict situations and other community-based humanitarian response efforts, to understand how they differ depending on local security conditions, access to infrastructure, and the way humanitarian response is organized. Longitudinal studies could follow the same responders over time to examine whether access to GenAI, confidence in using it, and local rules for its use change as the conflict, the technology, and their expertise evolve.

\section{Conclusion}
This work examines how community-based humanitarian responders in Sudan use, limit, and avoid GenAI during active armed conflict. The findings demonstrate that war can both increase the value of GenAI when human assistance is difficult to reach, while also weakening the conditions required for responsible use. In such cases, GenAI may provide provisional assistance, but its use still depends on current local knowledge, appropriate human verification, and whether accessing the technology is safe. Responsible use of GenAI in the context of ongoing conflict cannot be considered solely in terms of the performance of models or individual adoption, but must also account for the conditions under which humanitarian responders work. Supporting responsible use means supporting the judgment of responders who decide, often under threat, when these tools help and when they put people at risk.


\section{Generative AI Usage Statement}
We used GPT-6 to assist the native Arabic-speaking researcher with translating the Arabic-language interviews into English. The researcher manually reviewed the translations against the original Arabic to ensure accuracy. We also used GPT-6 alongside QCAmap to generate candidate codes, identify patterns for researcher review, and refine the grammar and wording of the manuscript. We reviewed all translations, coding decisions, and interpretations, and the final text was reviewed and approved by the authors.

\section{Ethical Considerations and Potential Adverse Impacts}
Research during active conflict involved risks of distress and indirect identification. We mitigated these through informed consent, de-identification, restricted data access, and review of potentially identifying contextual details. We also considered the possibility that publishing participants' accounts could expose sensitive humanitarian activities, so we generalized such details where necessary and had location and role descriptors reviewed by a research team member in the ERR network and an additional person familiar with it.
\bibliographystyle{ACM-Reference-Format}
\bibliography{sample-base}
\clearpage

\title{Supplementary Material}

\appendix

\section{Exploratory Survey}
\label{app:appendixA}

\makeatletter\if@twocolumn\onecolumn\fi\makeatother
\subsection{Exploratory Survey: Descriptive Results}\label{app:survey_results}
Before undertaking the survey, we provided the respondents with information about the research and a consent form. The term "GenAI" was explained as referring to AI generating content, and the examples mentioned included ChatGPT, Gemini, Copilot, and Claude. The resulting analytical sample consisted of 89 respondents. In this sample, 49 respondents (55.1\%) used GenAI in their ERR activities, 30 respondents (33.7\%) did not use it, and 10 respondents (11.2\%) were unsure. Among those who reported the use of GenAI, the most common activities included translation (22/49, 44.9\%), data analysis (21/49, 42.9\%), and creation of social media content (20/49, 40.8\%). Among those who reported no use of GenAI, the top three answers were lack of knowledge about how to use GenAI (8/30, 26.7\%), distrust of the output (7/30, 23.3\%), and unreliable internet connection (5/30, 16.7\%). These statistics reflect the participants recruited into the survey, and it is not safe to generalize them for all of the ERR members. The full questionnaire employed in the exploratory survey is provided below.

\begin{table*}[h]
\centering\small
\caption{Characteristics of exploratory survey respondents ($N=89$). Counts are based on each respondent's selected primary role, tenure and geographic area primarily supported. Regions are aggregated for confidentiality.}
\label{tab:err_survey_profile}
\begin{minipage}[t]{0.47\textwidth}
\centering\begin{tabular}{@{}lrr@{}}
\toprule
Characteristic & $n$ & \% \\ \midrule
\multicolumn{3}{@{}l}{\textit{Primary role}}\\
Coordinator/leader & 31 & 34.8 \\
Field volunteer & 17 & 19.1 \\
Communications/media & 11 & 12.4 \\
Data/reporting & 10 & 11.2 \\
Finance/administration & 7 & 7.9 \\
Other roles & 13 & 14.6 \\
\addlinespace\multicolumn{3}{@{}l}{\textit{Time working with ERRs}}\\
Less than 3 months & 2 & 2.2 \\
3--6 months & 13 & 14.6 \\
7--12 months & 5 & 5.6 \\
More than 1 year & 67 & 75.3 \\
Prefer not to answer & 2 & 2.2 \\
\bottomrule\end{tabular}
\end{minipage}\hfill
\begin{minipage}[t]{0.47\textwidth}
\centering\begin{tabular}{@{}lrr@{}}
\toprule
Characteristic & $n$ & \% \\ \midrule
\multicolumn{3}{@{}l}{\textit{Geographic area supported}}\\
Kordofan states & 31 & 34.8 \\
Darfur states & 31 & 34.8 \\
Khartoum & 12 & 13.5 \\
Other states & 12 & 13.5 \\
Prefer not to answer & 3 & 3.4 \\
\addlinespace\multicolumn{3}{@{}l}{\textit{GenAI use in ERR work}}\\
Yes & 49 & 55.1 \\
No & 30 & 33.7 \\
Not sure & 10 & 11.2 \\
\bottomrule\end{tabular}
\end{minipage}
\end{table*}

\begin{table}[h]
\centering\small
\caption{Reported tools and working conditions among respondents who indicated that they used GenAI in ERR work ($n=49$). Multiple tools and devices could be selected.}
\label{tab:err_survey_user_tools}
\begin{tabular}{@{}lrr@{}}\toprule
Characteristic & $n$ & \% \\ \midrule
\multicolumn{3}{@{}l}{\textit{Tools}}\\
ChatGPT & 32 & 65.3 \\
Gemini & 27 & 55.1 \\
Claude & 12 & 24.5 \\
Copilot & 4 & 8.2 \\
Name unknown & 6 & 12.2 \\
Other & 3 & 6.1 \\
\addlinespace\multicolumn{3}{@{}l}{\textit{Primary language used}}\\
Arabic & 26 & 53.1 \\
English & 1 & 2.0 \\
Both & 21 & 42.9 \\
Other & 1 & 2.0 \\
\addlinespace\multicolumn{3}{@{}l}{\textit{Devices}}\\
Personal phone & 43 & 87.8 \\
Shared phone & 2 & 4.1 \\
Personal computer/laptop & 16 & 32.7 \\
Shared computer/laptop & 4 & 8.2 \\
\bottomrule\end{tabular}
\end{table}

\begin{figure*}[h]
\centering
\begin{minipage}[t]{0.48\textwidth}\centering
\textbf{A. Reported GenAI tasks}\\[2pt]
\begin{tikzpicture}
\begin{axis}[
width=0.73\linewidth,height=6.0cm,
xbar,bar width=8pt, xmin=0, xmax=55,
y dir=reverse, ytick={0,1,2,3,4,5,6},
yticklabels={Translation, Data analysis, Social media content, Writing/editing reports, Summarizing, Planning/coordination, Other},
yticklabel style={font=\scriptsize,align=right},tick label style={font=\scriptsize},
xlabel={Respondents (\%)},xlabel style={font=\scriptsize},
axis x line*=bottom,axis y line*=left,
point meta=x, nodes near coords={\pgfmathprintnumber[fixed,precision=1]{\pgfplotspointmeta}\%},
every node near coord/.append style={font=\scriptsize,anchor=west,xshift=2pt},
enlarge y limits=0.13,xtick distance=10,
]
\addplot[fill=blue!65,draw=none] coordinates {
(44.9,0)
(42.9,1)
(40.8,2)
(38.8,3)
(38.8,4)
(38.8,5)
(4.1,6)
};
\end{axis}\end{tikzpicture}
\end{minipage}
\hfill
\begin{minipage}[t]{0.48\textwidth}\centering
\textbf{B. Reasons for non-use}\\[2pt]
\begin{tikzpicture}
\begin{axis}[
width=0.73\linewidth,height=6.0cm,
xbar,bar width=8pt, xmin=0, xmax=40,
y dir=reverse, ytick={0,1,2,3,4,5,6,7},
yticklabels={Do not know how,Distrust of outputs,Unstable internet,No team use,Unsuitable devices,Privacy/security concerns,No time to learn,Other},
yticklabel style={font=\scriptsize,align=right},tick label style={font=\scriptsize},
xlabel={Respondents (\%)},xlabel style={font=\scriptsize},
axis x line*=bottom,axis y line*=left,
point meta=x, nodes near coords={\pgfmathprintnumber[fixed,precision=1]{\pgfplotspointmeta}\%},
every node near coord/.append style={font=\scriptsize,anchor=west,xshift=2pt},
enlarge y limits=0.13,xtick distance=10,
]
\addplot[fill=gray!70,draw=none] coordinates {
(26.7,0)
(23.3,1)
(16.7,2)
(13.3,3)
(6.7,4)
(3.3,5)
(3.3,6)
(6.7,7)
};
\end{axis}\end{tikzpicture}
\end{minipage}
\caption{Selected tasks among respondents reporting GenAI use ($n=49$; panel A), and selected reasons for non-use among respondents reporting no GenAI use ($n=30$; panel B). Percentages are calculated using the respective group denominators (49 users and 30 non-users).}
\label{fig:err_survey_tasks_barriers}
\end{figure*}
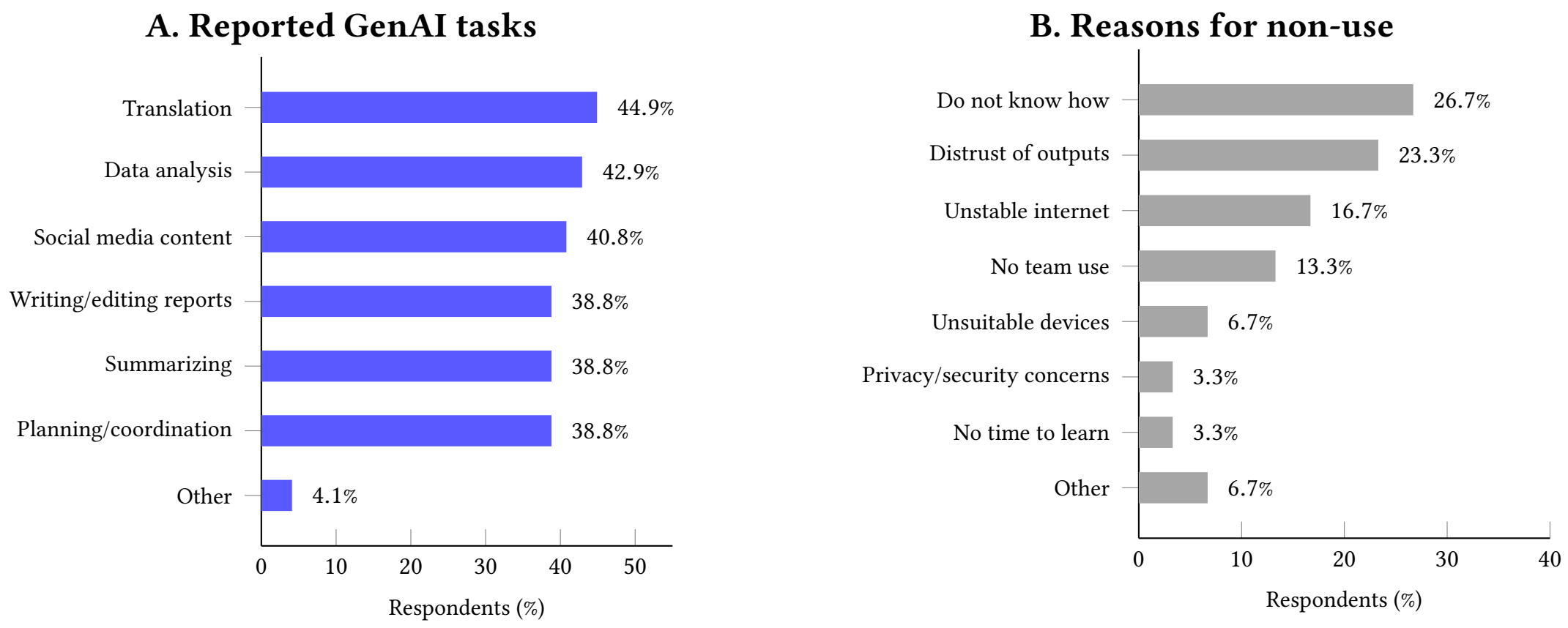

\begin{table*}[t]
\centering\footnotesize
\caption{Response distributions for all 23 Likert statements ($N=89$ per item). Values are percentages of respondents. Disagree combines strongly disagree and disagree; agree combines agree and strongly agree. Each item is stated as a claim or concern, so agreement is not uniformly favourable towards GenAI.}
\label{tab:err_survey_likert}
\begin{tabular}{@{}lp{0.67\textwidth}rrr@{}}
\toprule
Item & Statement (shortened) & Disagree & Neutral & Agree \\ \midrule
\addlinespace\multicolumn{5}{@{}l}{\textit{Technological factors}}\\
T1 & GenAI improves work quality or speed & 28.1 & 6.7 & 65.2 \\
T2 & Useful in fast-paced ERR work & 25.8 & 9.0 & 65.2 \\
T3 & Can use GenAI with little help & 31.5 & 15.7 & 52.8 \\
T4 & Supports different work tasks & 23.6 & 6.7 & 69.7 \\
T5 & Understands Sudanese local context & 33.7 & 16.9 & 49.4 \\
T6 & Reliable access to a device & 27.0 & 18.0 & 55.1 \\
T7 & Internet connection sufficiently stable & 38.2 & 16.9 & 44.9 \\
T8 & Can charge devices when needed & 39.3 & 13.5 & 47.2 \\
T9 & GenAI information is usually accurate & 34.8 & 23.6 & 41.6 \\
T10 & Arabic/local dialect outputs worse than English & 36.0 & 14.6 & 49.4 \\
\addlinespace\multicolumn{5}{@{}l}{\textit{Organisational factors}}\\
O1 & ERR leaders support GenAI use & 29.2 & 13.5 & 57.3 \\
O2 & ERR members open to new digital tools & 25.8 & 12.4 & 61.8 \\
O3 & ERRs can afford GenAI-related costs & 47.2 & 13.5 & 39.3 \\
O4 & Peers can help with GenAI use & 25.8 & 16.9 & 57.3 \\
O5 & Access to training or support & 29.2 & 22.5 & 48.3 \\
\addlinespace\multicolumn{5}{@{}l}{\textit{Environmental factors}}\\
E1 & Concern about exposing sensitive information & 28.1 & 21.3 & 50.6 \\
E2 & More funding would enable greater use & 22.5 & 10.1 & 67.4 \\
E3 & GenAI may weaken community trust & 37.1 & 31.5 & 31.5 \\
E4 & Government rules for GenAI are clear & 43.8 & 28.1 & 28.1 \\
E5 & Concern that inputs may be monitored & 30.3 & 20.2 & 49.4 \\
E6 & Usually need to verify GenAI outputs & 16.9 & 12.4 & 70.8 \\
E7 & Inaccurate outputs could harm beneficiaries & 29.2 & 20.2 & 50.6 \\
E8 & Human review needed in urgent situations & 21.3 & 15.7 & 62.9 \\
\bottomrule\end{tabular}
\end{table*}

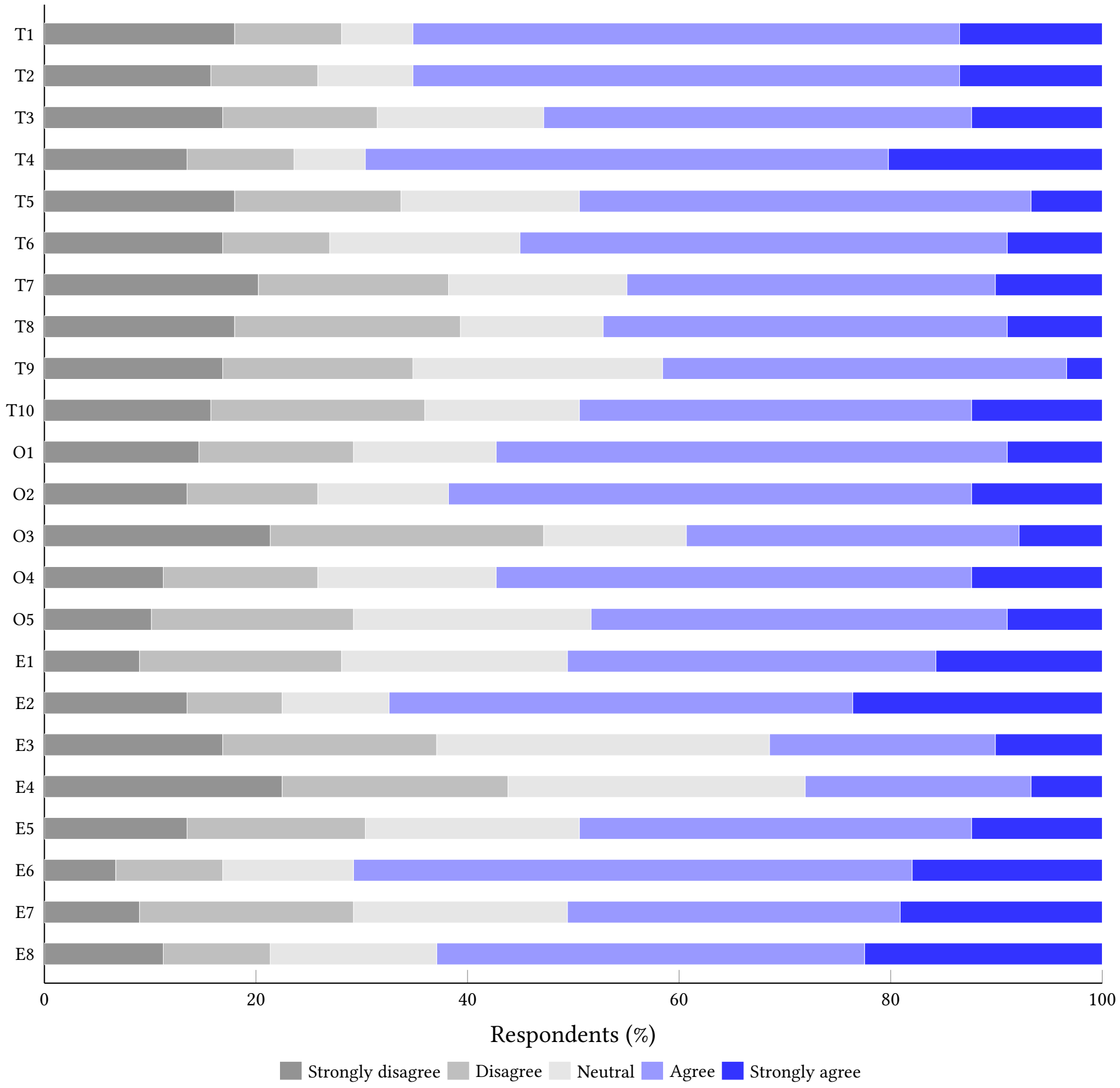
\begin{figure*}[t]
\centering
\begin{tikzpicture}
\begin{axis}[
width=0.90\textwidth,height=12.7cm,
xbar stacked,bar width=7pt,xmin=0,xmax=100,ymin=-0.7,ymax=22.7,
y dir=reverse, ytick={0,1,2,3,4,5,6,7,8,9,10,11,12,13,14,15,16,17,18,19,20,21,22},
yticklabels={T1,T2,T3,T4,T5,T6,T7,T8,T9,T10,O1,O2,O3,O4,O5,E1,E2,E3,E4,E5,E6,E7,E8},
yticklabel style={font=\scriptsize},tick label style={font=\scriptsize},
xlabel={Respondents (\%)},xlabel style={font=\small},
xtick={0,20,40,60,80,100},
axis x line*=bottom,axis y line*=left,legend style={at={(0.5,-0.07)},anchor=north,legend columns=5,font=\scriptsize,draw=none},
]
\addplot[draw=white,line width=0.15pt,fill=gray!85] coordinates {
(17.977528,0)
(15.730337,1)
(16.853933,2)
(13.483146,3)
(17.977528,4)
(16.853933,5)
(20.224719,6)
(17.977528,7)
(16.853933,8)
(15.730337,9)
(14.606742,10)
(13.483146,11)
(21.348315,12)
(11.235955,13)
(10.112360,14)
(8.988764,15)
(13.483146,16)
(16.853933,17)
(22.471910,18)
(13.483146,19)
(6.741573,20)
(8.988764,21)
(11.235955,22)
}; \addlegendentry{Strongly disagree}
\addplot[draw=white,line width=0.15pt,fill=gray!50] coordinates {
(10.112360,0)
(10.112360,1)
(14.606742,2)
(10.112360,3)
(15.730337,4)
(10.112360,5)
(17.977528,6)
(21.348315,7)
(17.977528,8)
(20.224719,9)
(14.606742,10)
(12.359551,11)
(25.842697,12)
(14.606742,13)
(19.101124,14)
(19.101124,15)
(8.988764,16)
(20.224719,17)
(21.348315,18)
(16.853933,19)
(10.112360,20)
(20.224719,21)
(10.112360,22)
}; \addlegendentry{Disagree}
\addplot[draw=white,line width=0.15pt,fill=gray!20] coordinates {
(6.741573,0)
(8.988764,1)
(15.730337,2)
(6.741573,3)
(16.853933,4)
(17.977528,5)
(16.853933,6)
(13.483146,7)
(23.595506,8)
(14.606742,9)
(13.483146,10)
(12.359551,11)
(13.483146,12)
(16.853933,13)
(22.471910,14)
(21.348315,15)
(10.112360,16)
(31.460674,17)
(28.089888,18)
(20.224719,19)
(12.359551,20)
(20.224719,21)
(15.730337,22)
}; \addlegendentry{Neutral}
\addplot[draw=white,line width=0.15pt,fill=blue!40] coordinates {
(51.685393,0)
(51.685393,1)
(40.449438,2)
(49.438202,3)
(42.696629,4)
(46.067416,5)
(34.831461,6)
(38.202247,7)
(38.202247,8)
(37.078652,9)
(48.314607,10)
(49.438202,11)
(31.460674,12)
(44.943820,13)
(39.325843,14)
(34.831461,15)
(43.820225,16)
(21.348315,17)
(21.348315,18)
(37.078652,19)
(52.808989,20)
(31.460674,21)
(40.449438,22)
}; \addlegendentry{Agree}
\addplot[draw=white,line width=0.15pt,fill=blue!80] coordinates {
(13.483146,0)
(13.483146,1)
(12.359551,2)
(20.224719,3)
(6.741573,4)
(8.988764,5)
(10.112360,6)
(8.988764,7)
(3.370787,8)
(12.359551,9)
(8.988764,10)
(12.359551,11)
(7.865169,12)
(12.359551,13)
(8.988764,14)
(15.730337,15)
(23.595506,16)
(10.112360,17)
(6.741573,18)
(12.359551,19)
(17.977528,20)
(19.101124,21)
(22.471910,22)
}; \addlegendentry{Strongly agree}
\end{axis}\end{tikzpicture}
\caption{Distribution of all 23 Likert responses ($N=89$). Labels T1--T10, O1--O5, and E1--E8 correspond to the technological, organizational, and environmental statements in Table~\ref{tab:err_survey_likert}. Items are not combined into composite scores.}
\label{fig:err_survey_likert}
\end{figure*}

\noindent\textit{Interpretation note.} Some respondents who reported not using GenAI in their ERR work also answered questions about GenAI tools and verification. We therefore report tasks, tools, and devices for the 49 self-reported users, and reasons for non-use for the 30 self-reported non-users. Likert results include all 89 respondents and measure reported agreement rather than actual behavior. Among these respondents, 16.9\% disagreed that GenAI outputs usually require verification, 21.3\% disagreed that human review is needed in urgent situations, and 29.2\% disagreed that inaccurate outputs could harm beneficiaries. We report descriptive results only, without inferential comparisons.

\begin{figure}[p]
    \centering
    \includegraphics[
        page=1,
        width=\linewidth
    ]{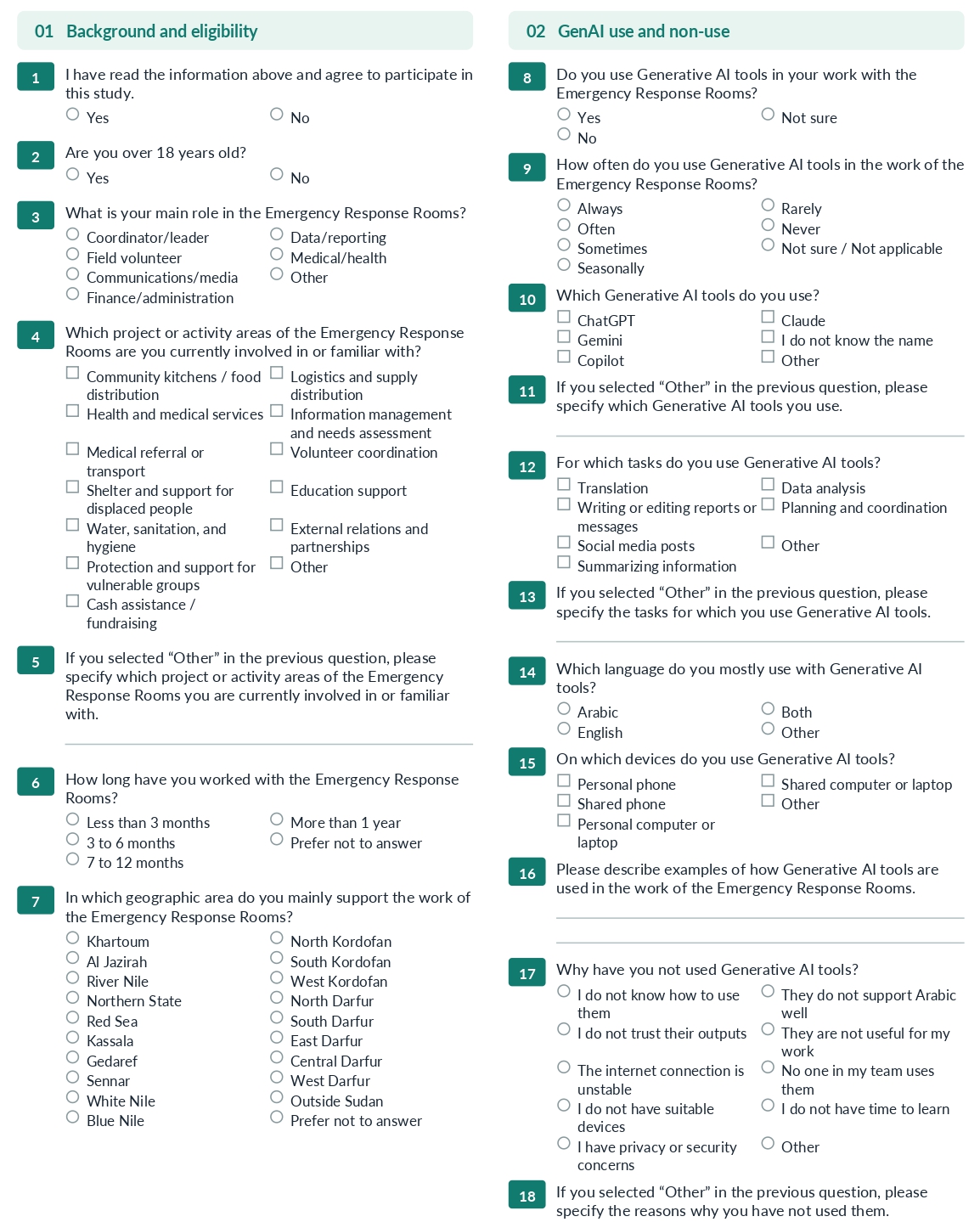}
    \caption{Exploratory survey questionnaire (Part 1).}
    \label{fig:SURVEY}
\end{figure}

\begin{figure}[p]
    \centering
    \includegraphics[
        page=2,
        width=\linewidth
    ]{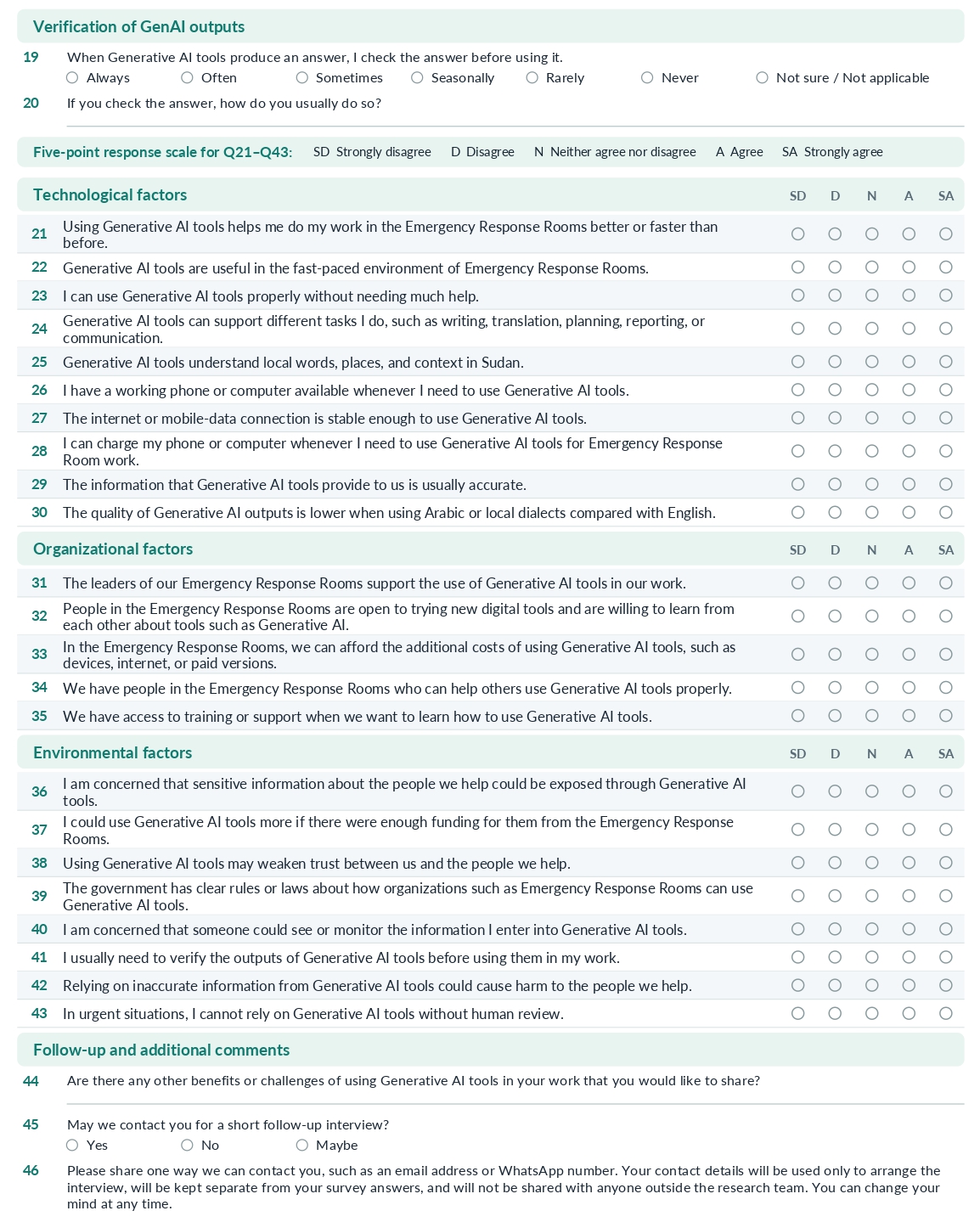}
    \caption{Exploratory survey questionnaire (Part 2).}
    
\end{figure}

\clearpage

\section{Interview Instrument, Coding and Thematic Analysis}
\label{app:appendixB}

\makeatletter\if@twocolumn\onecolumn\fi\makeatother

\subsection{Interview Instrument}
\label{app:interview-instrument}
\begin{figure}[h]
  \centering
  \includegraphics[width=\textwidth]{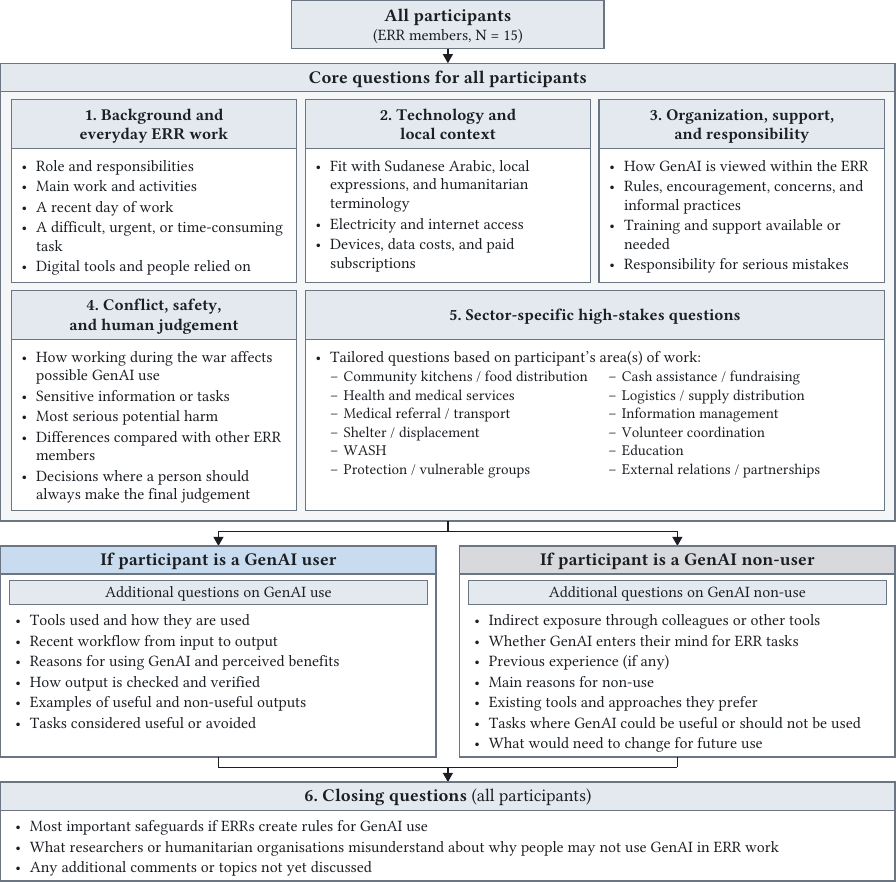}
  \caption{Overview of the semi-structured interview instrument ($N = 15$).}
  \Description{Flowchart of the interview guide. All participants answer core questions in five areas, then follow a user or non-user branch, and finish with closing questions.}
  \label{fig:interview-instrument}
\end{figure}

Interview recruitment sought variation across humanitarian roles and regions, although wartime insecurity and disrupted connectivity constrained participation from some areas. Figure~\ref{fig:interview-instrument} summarises the semi-structured interview guide, core questions for all participants, separate follow-up questions for GenAI users and non-users, and shared closing questions.

\subsection{Data Protection and Researcher Responsibility}

We conducted the study with the approval of the institutional ethics board. Before conducting the interviews, the interviewer informed each participant that anonymized transcripts would be translated and analyzed with the help of an AI model. The interviewees gave their verbal consent before continuing the process.

We only provided anonymized interview transcripts to GPT-6 using a paid ChatGPT Pro service with data sharing for model improvement turned off. Because LLMs can infer attributes such as place of residence from contextual details even after explicit identifiers are removed, we also generalized locations, roles, and other contextual details before submission.

We also note a tension between this practice and participants' own concerns about disclosing information to Generative AI (Section~4.3). We therefore limited model input to de-identified text and kept final analytical decisions with the researchers. The native Arabic-speaking researcher manually reviewed all GPT-6-assisted translations against the Arabic originals. Translation and coding nonetheless remain subject to interpretive limitations.

\subsection{Qualitative Coding Framework}\label{app:codebook}

The interview coding workbook contains 83 operational codes assigned to 217 selected meaning units from 15 interviews. The code IDs P01–P14 refer to operational codes, not to interview participants (P01–P15) in Section 4. The research team reviewed the codebook and confirmed the relationship between the coded material and the five final themes presented in Section 4. Table~\ref{tab:err_working_themes} presents the thematic groupings used during the analysis and their associated operational codes, while  Table~\ref{tab:err_complete_codebook}  provides the full code labels and coded-unit counts. These counts refer to coded passages, not the number of participants or the prevalence of an experience.

\setcounter{table}{0}
\renewcommand{\thetable}{\thesection\arabic{table}}
\begingroup\footnotesize\setlength{\LTpre}{6pt}\setlength{\LTpost}{6pt}\setlength{\tabcolsep}{5pt}
\begin{longtable}{@{}>{\raggedright\arraybackslash}p{0.49\linewidth}>{\raggedright\arraybackslash}p{0.44\linewidth}@{}}
\caption {Thematic groupings and associated operational code IDs used during the analysis.}\label{tab:err_working_themes}\\
\toprule \textbf{Working thematic group} & \textbf{Associated code IDs}\\ \midrule\endfirsthead
\multicolumn{2}{@{}l}{\textit{Table \thetable\ continued}}\\
\toprule \textbf{Working thematic group} & \textbf{Associated code IDs}\\ \midrule\endhead
\midrule \multicolumn{2}{r@{}}{\footnotesize Continued on next page}\\\endfoot
\bottomrule\endlastfoot
\addlinespace[5pt]\multicolumn{2}{@{}l}{\textbf{Practical benefits and value}}\\
Fallback when human support is disrupted & N01, N07, P06 \\
Sustaining care under constrained conditions & N02, N03, S11 \\
Augmentation while retaining human authority & P01, P07, P12 \\
Organisational translation and knowledge infrastructure & P03, P04, P14 \\
Administrative capacity under resource constraints & P06, P08, P12 \\
\addlinespace[5pt]\multicolumn{2}{@{}l}{\textbf{Adoption and access conditions}}\\
Fit with existing tasks and work gaps & P10, L01, T04, P02, P11, T01 \\
Learning through practical experience and peers & L02, L09, N04, T02 \\
Wartime digital access, costs and infrastructure & A01, A02, A05, A06, A08, A09, P05 \\
Unequal geographic and security conditions & A07, T05, T09, S03 \\
Emergency pressure and limits to reliance & N05, N06, N08, N09 \\
Organisational support and distributed knowledge & A04, A10, L10, P13 \\
Training and foundational capacity & L01, L03, L04, L05, L11 \\
Professional judgement and selective adoption & T01, T04, T07, S12 \\
\addlinespace[5pt]\multicolumn{2}{@{}l}{\textbf{Risks}}\\
Exposing people through humanitarian data & S01, S02, S07, S08, V08 \\
Physical exposure through digital traces & S03, T09 \\
Decisions made without reliable ground knowledge & E03, E08, V07 \\
Loss of local meaning in AI-mediated work & E02, E04, E05, E09 \\
Displacement of community knowledge and judgment & E01, E06 \\
Threats to dignity and legitimacy & S09, S10, E07 \\
Costs shifted onto frontline volunteers & A03, T08 \\
Network-wide security and governance gaps & S02, L06, L08 \\
Blurred responsibility for AI-assisted work & V04, V05 \\
\addlinespace[5pt]\multicolumn{2}{@{}l}{\textbf{Safeguards and human oversight}}\\
Keep sensitive information outside AI systems & S04, S05, S06 \\
Check outputs with evidence and human review & V01, V02, V03, V06, L12 \\
Co-design guidance with ERR volunteers & L07 \\
\end{longtable}\endgroup

\par\medskip
\noindent\textit{Coding note.} Some codes recur across working themes. The complete codebook below follows the workbook's analytic-role categories rather than assigning each code to a single final theme.
\begingroup
\footnotesize
\setlength{\LTpre}{6pt}\setlength{\LTpost}{6pt}
\setlength{\tabcolsep}{5pt}
\begin{longtable}{@{}p{0.10\linewidth}>{\raggedright\arraybackslash}p{0.75\linewidth}r@{}}
\caption{Operational codebook for the ERR interview analysis. Codes and passage counts are transcribed from the coding workbook.} \label{tab:err_complete_codebook}\\
\toprule
\textbf{Code} & \textbf{Operational code label} & \textbf{Units}\\
\midrule\endfirsthead
\multicolumn{3}{@{}l}{\textit{Table \thetable\ continued}}\\
\toprule
\textbf{Code} & \textbf{Operational code label} & \textbf{Units}\\
\midrule\endhead
\midrule\multicolumn{3}{r@{}}{\footnotesize Continued on next page}\\\endfoot
\bottomrule\endlastfoot
\addlinespace[4pt]
\multicolumn{3}{@{}l}{\textbf{Adoption and use processes} (48 codes)}\\
\addlinespace[1pt]
A01 & Losing digital time savings to unreliable power and connectivity & 3 \\
A02 & Prioritizing subsistence and aid over digital expenditure & 5 \\
A04 & Maintaining work through connected intermediaries and redundancy & 3 \\
A05 & Obtaining paid capacity through selective support arrangements & 6 \\
A06 & Finding that access barriers obstruct training itself & 3 \\
A07 & Anticipating conflict-driven geographic inequality in adoption & 3 \\
A08 & Facing payment, currency and platform restrictions & 1 \\
A09 & Requiring usable devices and infrastructure for AI participation & 5 \\
A10 & Negotiating institutional restrictions on humanitarian partnerships & 1 \\
L01 & Needing a practical entry point before usefulness becomes imaginable & 3 \\
L02 & Learning through experimentation and visible peer practice & 4 \\
L03 & Seeking training tailored to tasks and the conflict context & 5 \\
L04 & Combining sustained access with introductory learning & 1 \\
L05 & Prioritizing foundational capabilities before advanced AI use & 2 \\
L09 & Developing interest or proposed action through the interview & 2 \\
L10 & Recognizing limits of knowledge about other volunteers’ use & 2 \\
L11 & Seeking basic understanding of AI benefits, risks and operation & 5 \\
N01 & Turning to AI when danger makes other assistance inaccessible & 3 \\
N02 & Seeking partial action and hope rather than a complete solution & 1 \\
N03 & Extending humanitarian care beyond food through AI suggestions & 1 \\
N04 & Building practical trust through reported successful experience & 1 \\
N05 & Allowing necessity exceptions without endorsing routine reliance & 1 \\
N06 & Experiencing wartime urgency as pressure to adopt & 2 \\
N07 & Seeking AI support to interpret conflict-related threats & 1 \\
N08 & Requiring speed, flexibility and quality together in emergencies & 1 \\
N09 & Lacking time to use AI despite recognizing its value & 1 \\
P01 & Organizing human-supplied material while retaining authorship & 4 \\
P02 & Expanding AI from wording support into digital production & 4 \\
P03 & Turning meeting records into reusable organizational documentation & 1 \\
P04 & Translating across donor and volunteer language boundaries & 2 \\
P05 & Rationing limited AI capacity across tools or accounts & 4 \\
P06 & Replacing unavailable colleague time with on-demand assistance & 2 \\
P07 & Extending technical work while retaining specialist authority & 2 \\
P08 & Proposed AI-assisted digitization of beneficiary records & 2 \\
P09 & Using standard AI structures as scaffolding for local content & 1 \\
P10 & Leaving digital work with colleagues rather than personally adopting & 3 \\
P11 & Letting work demands and role changes shape AI use & 3 \\
P12 & Automating document conversion and comparison under human control & 1 \\
P13 & Coordinating practical work through existing tools and human expertise & 1 \\
P14 & Using AI-supported presentation in fundraising work & 1 \\
S11 & Defining wartime AI value through protection and basic needs & 1 \\
S12 & Anticipating continuing responsibilities beyond emergency response & 1 \\
T01 & Defining GenAI use by purpose rather than by tool name & 1 \\
T02 & Reading familiarity and interaction style as trust cues & 3 \\
T04 & Maintaining selective refusal despite professional capability & 3 \\
T05 & Changing adoption boundaries after a change in exposure & 1 \\
T06 & Questioning whom connected-user evidence represents & 1 \\
T07 & Judging AI by practical humanitarian value rather than output volume & 2 \\
\addlinespace[4pt]
\multicolumn{3}{@{}l}{\textbf{Risks} (21 codes)}\\
\addlinespace[1pt]
A03 & Risk \textemdash{} Digital coordination costs fall unevenly on frontline volunteers & 2 \\
E01 & Risk \textemdash{} AI-generated solutions can displace situated community knowledge & 8 \\
E02 & Risk \textemdash{} AI mediation can distort collectively agreed meanings and decisions & 3 \\
E03 & Risk \textemdash{} Generated reporting can detach claims from what actually happened on the ground & 4 \\
E04 & Risk \textemdash{} Local language and context gaps shift interpretive repair work onto volunteers & 7 \\
E05 & Risk \textemdash{} Formulaic AI language can flatten human voice and lived specificity & 2 \\
E06 & Risk \textemdash{} Habitual AI dependence can erode skills, judgment, and authorship & 7 \\
E07 & Risk \textemdash{} Donor suspicion of AI authorship can undermine document legitimacy & 1 \\
E08 & Risk \textemdash{} Rapidly changing field conditions can invalidate AI-supported judgments & 1 \\
E09 & Risk \textemdash{} Polished AI outputs can misfit local language, culture, and material realities & 3 \\
S01 & Risk \textemdash{} Uploading identifiers can expose beneficiaries and volunteers to financial or security harm & 6 \\
S02 & Risk \textemdash{} One person’s data exposure can compromise teams and interrupt aid delivery & 2 \\
S03 & Risk \textemdash{} Devices, connectivity, and digital traces can create physical exposure in conflict & 5 \\
S07 & Risk \textemdash{} Shared digital accounts concentrate sensitive data and access dependencies & 3 \\
S08 & Risk \textemdash{} Uncertain data exposure and unknown downstream misuse & 6 \\
S09 & Risk \textemdash{} AI-assisted documentation and synthetic media can violate dignity, consent, or identity & 3 \\
T03 & Risk \textemdash{} Benign intent and model refusals can create a false sense of safety & 2 \\
T08 & Risk \textemdash{} Provider incentives and resource costs can conflict with humanitarian priorities & 1 \\
T09 & Risk \textemdash{} Location-dependent access and context can make AI behavior uneven across settings & 1 \\
V07 & Risk \textemdash{} AI errors can create rework, delay, and service disruption & 3 \\
V08 & Risk \textemdash{} Stated data-protection boundaries can coexist with potentially conflicting upload practices & 3 \\
\addlinespace[4pt]
\multicolumn{3}{@{}l}{\textbf{Safeguards} (8 codes)}\\
\addlinespace[1pt]
L12 & Safeguard \textemdash{} AI outputs remain drafts until explicit human review and approval & 4 \\
S04 & Safeguard \textemdash{} Use paper when digital collection threatens safety or community trust & 1 \\
S05 & Safeguard \textemdash{} Separate organizational visibility from individual identifiability & 1 \\
S06 & Safeguard \textemdash{} Keep sensitive operational records outside AI-assisted drafting & 1 \\
V01 & Safeguard \textemdash{} Verify AI outputs against known inputs and independent evidence & 4 \\
V02 & Safeguard \textemdash{} Increase verification as technical and humanitarian stakes rise & 2 \\
V03 & Safeguard \textemdash{} Route AI-assisted outputs through collective human review workflows & 2 \\
V06 & Safeguard \textemdash{} Cross-model checking still requires human verification & 2 \\
\addlinespace[4pt]
\multicolumn{3}{@{}l}{\textbf{Accountability} (2 codes)}\\
\addlinespace[1pt]
V04 & Accountability \textemdash{} The submitter or designated role remains responsible for AI-assisted work & 3 \\
V05 & Accountability \textemdash{} Collective review distributes responsibility before release & 4 \\
\addlinespace[4pt]
\multicolumn{3}{@{}l}{\textbf{Governance risks} (1 code)}\\
\addlinespace[1pt]
L06 & Governance risk \textemdash{} Informal AI use across an open volunteer network can outpace safeguards & 3 \\
\addlinespace[4pt]
\multicolumn{3}{@{}l}{\textbf{Governance safeguards} (1 code)}\\
\addlinespace[1pt]
L07 & Governance safeguard \textemdash{} Co-design AI guidance with volunteers to fit field realities & 2 \\
\addlinespace[4pt]
\multicolumn{3}{@{}l}{\textbf{Governance gaps} (1 code)}\\
\addlinespace[1pt]
L08 & Governance gap \textemdash{} Existing policies can leave AI-specific risks unaddressed & 2 \\
\addlinespace[4pt]
\multicolumn{3}{@{}l}{\textbf{Boundaries of acceptable use} (1 code)}\\
\addlinespace[1pt]
S10 & Boundary \textemdash{} Humanitarian protection and dignity constrain acceptable AI use & 3 \\
\end{longtable}
\endgroup
\noindent\textit{Note.} Units refer to coded passages, not unique participants. Code frequencies were not used to determine thematic importance or select final themes. The code IDs P01–P14 refer to operational codes, not to interview participants (P01–P15) in Section 4.

\subsection{Development of Final Themes}

The 83 operational codes were organized into broader thematic groupings during the collaborative analysis workshop. These groupings were not mapped one-to-one to the five final themes. Instead, the research team examined relationships and overlaps between codes, compared experiences across participants, and revisited the interview transcripts to develop the five themes presented in Section 4. Twenty codes were not mapped to a final theme; they remain in the codebook but did not form part of the five themes reported in Section 4.

\begin{table}[htbp]
\centering
\small
\caption{Mapping of operational codes to the five final themes.}
\label{tab:final-theme-mapping}

\begin{tabularx}{\linewidth}{@{}p{0.43\linewidth}X@{}}
\toprule
\textbf{Final theme} &
\textbf{Relevant operational codes} \\
\midrule

4.1 Human support hard to reach &
N01, N02, N03, N04, N05, N06, N07, N08,
P06, P07, S11 \\

\addlinespace

4.2 Wartime ground truth &
E01, E04, E06, E08, E09, P01, P07,
T07, V01, V02, V06 \\

\addlinespace

4.3 Protection concerns &
S01, S02, S03, S04, S05, S06, S07,
S08, S10, T03, V08 \\

\addlinespace

4.4 Humanitarian communication credibility &
P01, P04, P14, E02, E03, E04, E05,
E07, E09, S09 \\

\addlinespace

4.5 Unequal access and collective use &
A01, A02, A03, A04, A05, A06, A07,
A08, A09, L01, L02, L03, L05, P05,
P10, P11, P13, S07, T04, T05, T06,
T09, V03, V04, V05 \\

\bottomrule
\end{tabularx}
\end{table}

\end{document}